\documentclass[letterpaper, 10 pt, conference]{ieeeconf}
\usepackage{booktabs}
\IEEEoverridecommandlockouts                              % This command is only needed if 
\usepackage{graphics} % for pdf, bitmapped graphics files
\usepackage{epsfig} % for postscript graphics files
\usepackage{mathptmx} % assumes new font selection scheme installed
\usepackage{times} % assumes new font selection scheme installed
\usepackage{amsmath} % assumes amsmath package installed
\usepackage{amssymb}  % assumes amsmath package installed
\usepackage[ruled,vlined,linesnumbered]{algorithm2e}
\usepackage[urlcolor  = blue, colorlinks = true]{hyperref}
\usepackage[table,xcdraw]{xcolor}
\usepackage{cite}
\usepackage{float}
\usepackage{algpseudocode}

\usepackage{booktabs}
\usepackage{graphicx}
\usepackage[table,xcdraw]{xcolor}

\let\oldnl\nl% Store \nl in \oldnlRemove line number
\newcommand{\nonl}{\renewcommand{\nl}{\let\nl\oldnl}}%

\title{\LARGE \bf
Multiplicative Controller Fusion: Leveraging Algorithmic Priors for Sample-efficient Reinforcement Learning and Safe Sim-To-Real Transfer
}

\author{Krishan Rana, Vibhavari Dasagi, Ben Talbot,  Michael Milford, and Niko S\"{u}nderhauf% <-this % stops a space
\thanks{This work was conducted by the Australian Research Council Centre of Excellence for Robotic Vision under project CE140100016 and supported by the QUT Centre for Robotics. The authors are with the Australian Centre for Robotic Vision at Queensland University of Technology (QUT),
Brisbane, QLD 4001, Australia.
        {\tt\small krishan.rana@hdr.qut.edu.au}}%
}

\begin{document}

\maketitle
\thispagestyle{empty}
\pagestyle{empty}
%%%%%%%%%%%%%%%%%%%%%%%%%%%%%%%%%%%%%%%%%%%%%%%%%%%%%%%%%%%%%%%%%%%%%%%%%%%%%%%%
\begin{abstract}

Learning-based approaches often outperform hand-coded algorithmic solutions for many problems in robotics. However, learning long-horizon tasks on real robot hardware can be intractable, and transferring a learned policy from simulation to reality is still extremely challenging. We present a novel approach to model-free reinforcement learning that can leverage existing sub-optimal solutions as an \emph{algorithmic prior} during training and deployment. During training, our gated fusion approach enables the prior to guide the initial stages of exploration, increasing sample-efficiency and enabling learning from sparse long-horizon reward signals. Importantly, the policy can learn to improve beyond the performance of the sub-optimal prior since the prior's influence is annealed gradually. During deployment, the policy's uncertainty provides a reliable strategy for transferring a simulation-trained policy to the real world by falling back to the prior controller in uncertain states. We show the efficacy of our Multiplicative Controller Fusion approach on the task of robot navigation and demonstrate safe transfer from simulation to the real world without any fine-tuning.
The code for this project is made publicly available at \url{https://sites.google.com/view/mcf-nav/home}

\end{abstract}

\section{Introduction}
Deep reinforcement learning (RL) shows immense potential for autonomous navigation agents to learn complex behaviours that are typically difficult to specify analytically via classical, hand-crafted approaches. However, they often require extensive amounts of online training data, a limiting factor for real-world robot applications. Additionally, RL policies overfit to their training environment, making them unreliable for safe deployment in new environments, particularly when transferring from simulation to the real world. On the other hand, classical approaches for reactive navigation can guarantee safety and can reliably adapt to diverse environments via parameter-tuning. They, however, lack the capabilities to efficiently navigate in cluttered environments, are susceptible to oscillations\cite{khatib1986real}, and seizure in local minima \cite{koren1991potential}. The high-level skills required to overcome these inefficiencies are difficult to hand-engineer explicitly and tend to deteriorate in performance when extensively tuned for a particular environment.

\begin{figure}[thpb]
  \centering
  \includegraphics[width=0.49\textwidth]{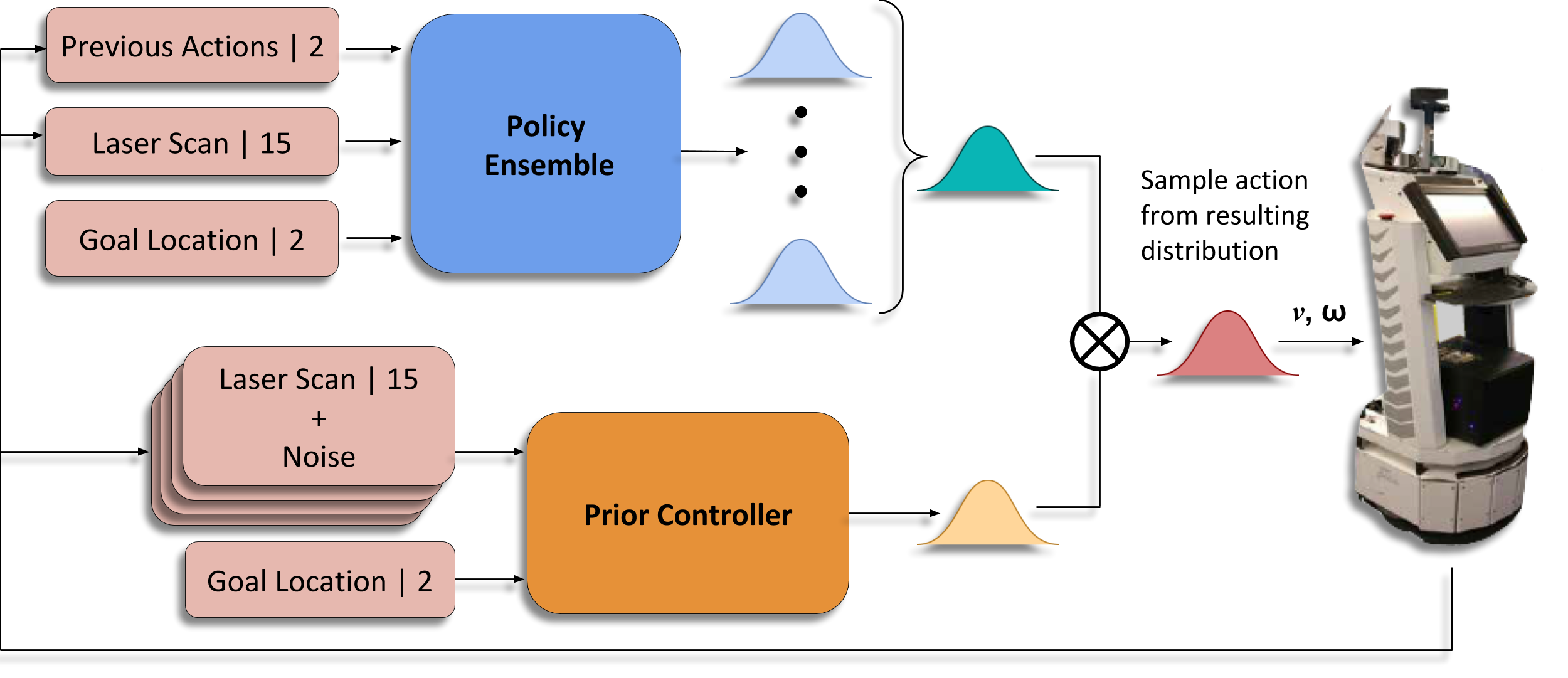}
  \vspace{-0.5cm}
  \caption{Multiplicative Controller Fusion (deployment) system diagram for real world navigation. Policy ensemble shown was trained in simulation. In cases where policy ensemble exhibits high uncertainty in the real world, the resulting distribution will bias towards the prior controller distribution allowing for safer navigation. }
  \label{front}
  \vspace{-0.47cm}
\end{figure}

In this work, we combine classical and learned strategies in order to address their respective limitations. We present Multiplicative Controller Fusion (MCF), an approximate Bayesian approach which fuses a classical controller (prior) and stochastic RL policies for guided policy exploration during training, and safe navigation during deployment (see Figure \ref{front}). This work primarily focuses on a goal-directed navigation task with sparse rewards, but can be extended to any continuous control task where a competent prior is available.\\
As opposed to Reinforcement Learning from Demonstrations (RLfD), MCF operates directly within the exploration phase and does not require an expert prior controller. It additionally utilises a single objective to optimise the policy. We formulate the action output from the prior as a distribution around the deterministic action. Sampling from this distribution during the early stages of training allows the agent to explore the most relevant regions for the task including the surrounding state-action pairs, which stabilises training. As the agent's performance improves, a gating function gradually shifts the resulting sampling distribution towards the policy, allowing the agent to fully exploit its behaviours, and improve beyond that of the prior. At deployment, MCF provides an uncertainty-aware navigation strategy for safe deployment in the real world. The multiplicative fusion of the two distributions results in a composite distribution which naturally biases towards the controller exhibiting the least uncertainty at a given state while attaining the performance of the learned system where it is more confident.

To summarise, the key contributions of our paper are:

\begin{itemize}
    \item a novel training strategy which utilises a sub-optimal controller to guide exploration for continuous control tasks involving sparse, long-horizon rewards, demonstrating significant improvements to the sample efficiency and the ability for the agent to improve beyond the performance of the controller;
    \item a novel deployment strategy for continuous control reactive navigation agents which leverages policy uncertainty estimates to safely fall back to a risk-averse prior controller in states of high policy uncertainty;
    \item an evaluation of our approach to allow a simulation trained policy to transfer to the real world for safe navigation in cluttered environments without any fine-tuning, capable of outperforming both the prior and end-to-end trained systems. 
\end{itemize}

\section{Related Work}

\subsection{Classical Navigation Approaches}
Classical approaches to navigation can be largely divided into two categories: \textit{deliberative} and \textit{reactive} systems. Deliberative systems typically rely on the availability of a globally consistent map to plan out safe trajectories \cite{Cadena_2016}, whereas reactive systems rely on the immediate perception of their surrounding environment. This allows them to handle dynamic objects and those unaccounted for in the global map. Vector Field Histograms (VHF) \cite{borenstein1991vector} is a real-time motion planning algorithm that generates a polar histogram to represent the polar density of surrounding obstacles. The robot's steering angle is then chosen based on the direction exhibiting the least density and closeness to the goal. For this approach, the polar histogram has to be computed at every time step, making it suitable for dynamic obstacle avoidance. Artificial Potential Field (APF) \cite{warren1989global, hwang1992potential} based approaches compute a local potential function which attracts the robot towards the goal while repelling it away from obstacles. Other approaches leverage short term memory \cite{antich2005extending} in order to build local maps of the surrounding environment, allowing the agent to identify potential dead-ends. The bug family of algorithms \cite{ng2007performance} can guarantee completeness when searching for a goal but lack efficiency.\\ 
A common disadvantage of all these approaches is the need for extensive tuning and hand engineering to achieve good performance with a tendency to deteriorate in performance when tuned for a particular domain. Additionally, they are susceptible to oscillations, getting stuck in local minima \cite{koren1991potential,khatib1986real} and exhibit suboptimal path efficiency.

\subsection{Learning for Reactive Navigation}
Learning navigation strategies is an attractive alternative to robot control when compared to classical analytically derived approaches. Common approaches are imitation learning \cite{argall2009survey} and deep reinforcement learning \cite{mnih2015human}. Imitation learning trains a model by mimicking a set of demonstrations provided by an expert. Such approaches have been shown to successfully teach an agent to navigate along forest trails \cite{ross2013learning}, and accomplish uncertainty aware visual navigation tasks \cite{yunvisual}. Kim \textit{et al.} \cite{kim2015deep} and Pfeifer \textit{et al.} \cite{pfeiffer2017perception} train a navigation agent using a global path planner as the set of labelled demonstration data. These systems are however limited by the performance of the demonstration set. On the other hand, deep reinforcement learning based approaches do not rely on a dataset and rather allow the agent to gain experience via interaction with the environment. Zhu \textit{et al.} \cite{zhu2017target} train a monocular based robot for target driven navigation in a high fidelity simulation environment and fine-tune it for deployment in the real world. Given the close correspondence between laser scans in simulation and the real world, Tai \textit{et al.}\cite{tai2017virtual} and Xie \textit{et al.}\cite{xie2018learning} show that training an agent in simulation can be transferred directly to a real robot without any fine-tuning when using laser-based sensors. Despite showing reasonable performance, the robot was still shown to fail in scenarios it did not generalise to during training. Recent approaches have made attempts to improve the safety of these systems when presented with unknown states. Kahn \textit{et al.} \cite{kahn2017uncertainty} propose an uncertainty-aware navigation strategy using model-based learning. They rely on uncertainty estimates of a collision prediction module and utilise this as a risk term in model predictive control (MPC). Lotjens \textit{et al.} \cite{lutjens2018safe} extend this approach to avoid dynamic obstacles in complex scenarios by utilising an ensemble of LSTM networks to estimate the uncertainty of surrounding obstacles. We extend these ideas to model-free reinforcement learning and propose a unified approach which leverages risk-averse prior controllers for safe real-world deployment.

\subsection{Combining Classical and Learned Systems}
Instead of assuming no prior knowledge, a better-informed alternative is to leverage the large body of classical approaches to aid learning-based systems. Several recent works have taken steps towards this notion. Xie \textit{et al.} \cite{xie2018learning} leverage a proportional controller to speed up the training process during exploration in navigation tasks. The idea is based on the hand-crafted controller yielding higher rewards than random exploration alone. Bansal \textit{et al.} \cite{bansal2019combining} train a perception module to produce obstacle-free waypoints with which an optimal controller can path plan towards. Our prior work \cite{rana2019residual} proposes a tightly coupled training approach between a classical controller and reinforcement learning, based on the residual reinforcement learning framework. A residual policy is trained to improve the performance of a suboptimal classical controller while leveraging the classical controller to guide the exploration. Similarly, Iscen \textit{et al.} \cite{Iscen2018PoliciesMT} demonstrate how a learned policy can be used to modify trajectory generators to improve their base level of performance and show its applicability for real robot locomotion. These approaches however only learn a residual policy to modify the prior, limiting the expressiveness of the overall system. We formulate an alternative approach which gradually allows the policy to be independent of the classical controller enabling it to improve far beyond its performance. Reinforcement Learning from Demonstrations (RLfD) \cite{hester2018deep, rajeswaran2017learning, vecerik2017leveraging, gao2018reinforcement} provides an alternative approach to introducing prior knowledge from classical systems to aid the learning process, however heavily rely on the presence of perfect demonstrations and utilise multiple objectives to optimise the system in order to stabilise training. Our approach does not rely on perfect demonstrations and can instead leverage demonstrations from suboptimal handcrafted controllers to guide the learning process, gradually allowing the policy to improve beyond their inefficiencies. We additionally leverage these classical controllers as a safe fallback in cases of high policy uncertainty when deployed in the real world. 

\section{Problem Formulation}
We consider the reinforcement learning framework in which an agent learns an optimal controller for a given task through environment interaction. While providing an attractive solution to solve complex control tasks which are difficult to derive analytically, their application to real robots is plagued by high sample inefficiency during training and unsafe behaviour in unknown states. This hinders the overall adoption of these systems in the real world. We propose an approach which leverages the vast number of suboptimal classical controllers (priors) developed by the robotics community to address these limitations.

Traditionally in RL, an agent begins by randomly exploring its environment, starting at any given state $s$, the agent performs an action $a$ and arrives in state $s'$, receiving a reward $r(s,a,s')$ $\in$ $\mathbb{R}$. In order to learn an optimal policy, the goal of the agent is to maximise the expectation of the sum of discounted rewards, known as the return $R_{t} = \sum^{\infty}_{i=t} \gamma^{i}r(s_{i},a_{i},s_{i+1})$, which weighs future rewards according to the discount factor $\gamma$. In this work, we leverage existing classical controllers to both guide exploration during training as well as provide safety guarantees during deployment. We assume that a suboptimal classical controller is present for this task and that its explicit analytical derivation can provide such performance guarantees, which generally are not provided by learned policies. We refer to this as risk-averse behaviour.

\section{Multiplicative Controller Fusion}
\label{sec:mcf}
We introduce Multiplicative Controller Fusion (MCF), which takes a step towards unifying classical controllers and learned systems in order to attain the best of both worlds during training and deployment. Our approach focuses on continuous control tasks, a staple in robotics. We formulate the action outputs from both the prior controller and policy as distributions over actions where a composite policy is obtained via a multiplicative composition of these distributions. The general form of our approach is given by:

\begin{equation}
\label{eq1}
    \pi'(a|s) = \frac{1}{Z}(\pi_{\theta}(a|s)\cdot\pi_{prior}(a|s))  
\end{equation}

where $\pi_{\theta}(a|s)$ and $\pi_{prior}(a|s)$ represents the distribution over actions from the policy network and prior controller respectively. $Z$ is a normalisation coefficient which ensures that the composite distribution $\pi'(a|s)$ is normalised. 

\subsection{Components}
MCF consists of two components: a reinforcement learning policy and a classical controller which we refer to as the prior. We describe each of these systems below.

\subsubsection*{\textbf{Policy}}
 We leverage stochastic RL algorithms that output each action as an independent Gaussian $\pi_{\theta}(a|s) = \mathcal{N}(\mu, \Sigma)$ with $\mu$ containing the component-wise means $\mu_{v}$ for the linear velocity and $\mu_{\omega}$ for angular velocity. The diagonal covariance matrix $\Sigma$ contains the corresponding variances $\sigma^{2}_{v}$ for the linear velocity and $\sigma^{2}_{\omega}$ for the angular velocity. This output distribution is suitable for use during training, however since the distribution is trained to maximise entropy over the actions, its variance does not reflect the model's uncertainty. At deployment it is particularly important that the distribution provides a representation of the policy's state uncertainty, known as \textit{epistemic} uncertainty. To attain such a distribution at deployment, we follow approaches for uncertainty estimation from the deep learning literature based on training ensembles \cite{chua2018deep}. \textit{N} randomly initialised policies are trained and at inference, the agreement between them at a given state indicates the level of uncertainty. We employ this idea at the deployment phase on a real robot.

\subsubsection*{\textbf{Prior Controller}}
We utilise the large body of work developed by the robotics community consisting of analytically derived controllers. These hand-crafted controllers demonstrate competent levels of performance across various domains but are inefficient in unstructured environments and are limited to the behaviours defined at the design stage. For uncertainty-aware deployment, MCF relies on a distribution over the actions. However, most classical methods do not produce principled and calibrated uncertainty estimates. Therefore, we construct an approximate distribution based on the sensor noise, which is the main source of uncertainty in these systems. We do this by propagating the sensor uncertainty using Monte Carlo sampling to extract an uncertainty estimate over the action space. For guided exploration, the training distribution primarily serves as a medium for Gaussian exploration, allowing the policy to explore the surrounding state-action pairs for potential improvements.

\subsection{Guided Exploration}
Exploration is difficult in sparse long horizon settings for standard reinforcement learning techniques, requiring large amounts of environment interaction. As an alternative to random Gaussian exploration, we impose a soft constraint on exploration by guiding it during the initial stages of training using a prior controller.
We utilise a gated variant of Equation \ref{eq1} for Gaussian exploration using the composite distribution. The gating function biases the composite distribution $\pi'(a|s)$ towards the prior early on during training, exposing it to the most relevant parts of the state-action space. As the policy becomes more capable, the gating function gradually shifts $\pi'(a|s)$ towards the policy distribution by the end of training. This allows the policy to fully exploit its learned behaviours and improve beyond the prior. The multiplicative fusion constrains the exploration such that the policy does not deviate far off from the prior, exploiting unwanted behaviours.

\begin{equation}
\label{eq:gated}
    \pi'(a|s) = \frac{1}{Z}(\pi_{\theta}(a|s)^{1-\alpha}\cdot\pi_{prior}(a|s)^{\alpha})
\end{equation}

Equation \ref{eq:gated} defines the gated form of our multiplicative fusion strategy used during training, where $\alpha$ represents the gating term that shifts the resulting distribution. By formulating the action outputs of the prior as a unimodal Gaussian, we allow the surrounding state-action regions of the Q-value network to be correctly updated, reducing the chances of overestimation bias and stabilising training. The gating formulation is reminiscent of the \textit{epsilon-greedy} strategy used in Q-learning \cite{mnih2015human}, however, as opposed to totally random actions, we utilise a distribution around a competent controller. The complete MCF Training algorithm is shown in Algorithm \ref{algorithm1}.

\begin{algorithm}[t]
\label{algorithm1}
\SetAlgoLined
\textbf{Given:} Given Policy Network $\pi_{\theta}$, Prior Controller $\pi_{prior}$ \\
\KwIn{State $\textbf{\textit{s}}_{t}$, Gating Function \textbf{$\alpha$}}
\KwOut{Trained Policy,  $\pi_{\theta}$}
\For{$t=1$ \KwTo $T$}{
   Compute the composite distribution $\pi'(a|s_{t}) \sim \mathcal{N}(\mu', \sigma^{2'})$\\
  
  \nonl \hspace{0.5cm} $\pi'(a|s_{t}) = \frac{1}{Z}(\pi_{\theta} (a|s_{t})^{(1-\alpha)}\cdot\pi_{prior}(a|s_{t})^{\alpha})$\\
  
  Sample action \textbf{\textit{a}} from the new distribution $\pi'(a|s_{t})$ and step in environment.\\
  
  Store $(s_{t}, a, r, s_{t+1})$\\
  
  Update $\pi_{\theta}$\\
  
  Compute $\alpha$\\
}

 \Return {$\pi_{\theta}$}
 \caption{MCF Training}
\end{algorithm}

The mean $\mu'$ and variance $\sigma^{2'}$ defining the composite Gaussian $\pi'(a|s_{t})$ can be expressed as follows,

\begin{equation}
    \mu' = \frac{\mu_{\theta}\sigma_{prior}^{2}(1-\alpha) + \mu_{prior}\sigma_{\theta}^{2}\alpha}{\sigma_{prior}^{2}(1-\alpha) + \sigma_{\theta}^{2}\alpha}
\end{equation}

\begin{equation}
    \sigma^{2'} = \frac{\sigma^{2}_{\theta}\sigma_{prior}^{2}}{\sigma_{prior}^{2}(1-\alpha) + \sigma_{\theta}^{2}\alpha}
\end{equation}

This expansion implicitly handles the normalisation term $\frac{1}{Z}$. The multiplicative formulation constrains the exploration process, allowing the agent to focus on the most relevant regions in the environment, reducing the overall training time. The gating parameter, $\alpha$ initially begins at 1, indicating a complete shift towards the prior distribution and gradually shifts entirely towards the policy distribution when its value is 0. The gating function should ideally be a function of the policy's performance during training however we leave the exploration of this idea to future work. In this work, we represent the gating function as a reverse logistic function which is a function of the training steps taken. \\
An advantage of using this type of fusion during exploration is that the policy distribution is biased towards the state-action trajectories followed by the prior. This mitigates the exploitation of unwanted behaviours commonly seen during random exploration. It also makes the policy suitable for multiplicative combination with the prior during deployment, as we will describe in the following section.

\subsection{Uncertainty-Aware Deployment}

At deployment, we directly utilise Equation \ref{eq1} to derive a composite policy, $\pi'(a|s)$ which demonstrates the complex navigational skills attained by the learned system while exhibiting the risk-averse behaviours of the prior in states of high policy uncertainty. This allows for efficient and safe real-world deployment. 
In order to attain this behaviour in the fusion process, we require the policy distribution to represent its state uncertainty. Given that all training is completed in a non-exhaustive simulation environment, the policy is bound to encounter states it did not generalise well to when transferred to the real world. We encapsulate this uncertainty by training an ensemble using Algorithm \ref{algorithm1} and computing the mean and variance of the action produced by the ensemble at a given state during inference. This allows the distribution to indicate a higher standard deviation in unknown states and lower values when all policies agree at familiar states. The complete MCF Deployment algorithm is described in Algorithm \ref{algorithm2}.

\begin{algorithm}[t]
\label{algorithm2}
\SetAlgoLined
\textbf{Given:} Ensemble of \textit{N} Trained Policies ($[\pi_{\theta 1}, \pi_{\theta 2} ... \pi_{\theta N}]$), Prior Controller ($\pi_{prior}$) \\
\KwIn{State $\textbf{\textit{s}}_{t}$}
\KwOut{Action \textbf{\textit{a}}}
  Approximate ensemble predictions as a unimodal Gaussian $\pi_{\theta}^{*}(a|s_{t}) = \mathcal{N}(\mu_{\theta}^{*}, \sigma_{\theta}^{2*})$, where:\\
  
  \nonl \hspace{0.5cm} $\mu_{\theta}^{*} = \frac{1}{N} \sum_{n=1}^{N} \mu_{\theta n}$\\
  
  \nonl \hspace{0.5cm} $\sigma_{\theta}^{2*} = \mathrm{Var}[\mu_{\theta n}]$\\  
  
  Compute the composite distribution $\pi'(a|s_{t}) \sim \mathcal{N}(\mu', \sigma^{2'})$\\
  
  \nonl \hspace{0.5cm} $\pi'(a|s_{t}) = \frac{1}{Z}(\pi_{\theta}^{*} (a|s_{t})\cdot\pi_{prior}(a|s_{t}))$\\
  
  Sample action \textbf{\textit{a}} from the distribution $\pi'(a|s_{t})$\\
  
 \Return {\textbf{\textit{a}}}
 \caption{MCF Deployment}
\end{algorithm}

The mean $\mu'$ and variance $\sigma^{2'}$ representing this composite Gaussian can be expressed as follows,

\begin{equation}
   \mu' = \frac{\mu^{*}_{\theta}\sigma_{prior}^{2} + \mu_{prior}\sigma_{\theta}^{2*}}{\sigma_{prior}^{2} + \sigma_{\theta}^{2*}} 
\end{equation}

\begin{equation}
    \sigma^{2'} = \frac{\sigma^{2*}_{\theta}\sigma_{prior}^{2}}{\sigma_{prior}^{2} + \sigma_{\theta}^{2*}}  
\end{equation}

where this expansion implicitly handles the normalisation term $\frac{1}{Z}$.

\section{Experiments}

In this work we focus on the goal-directed navigation task presented by Anderson \textit{et al.} \cite{anderson2018evaluation} which requires the agent to efficiently navigate to the desired goal while avoiding obstacles and dead-ends.

\subsection{Experimental Setup}
\subsubsection*{\textbf{Training Environment}}
All policy training was conducted in simulation and deployed in the real world without any additional fine-tuning. We utilise the laser-based navigation simulation environment provided in \cite{rana2019residual} to train all agents and transfer the trained policy to an identical robot in the real world. The training environment consists of 5 arenas with different configurations of obstacles. The goal and start location of the robot are randomised at the start of every episode, each placed on the extreme opposite ends of the arena (see Figure~\ref{explore}). This sets the long horizon nature of the task. As we focus on the sparse reward setting, we define $r(s,a,s') = 1$ if $d_{target} < d_{threshold} $ and $r(s,a,s') = 0$ otherwise, where $d_{target}$ is the distance between the agent and the goal and $d_{threshold}$ is a set threshold. The length of each episode is set to a maximum of 500 steps. The action space consists of two continuous values: linear velocity $v\in[-1,1]$ and angular velocity $\omega\in[-1,1]$. We assume that the robot can localise itself within a global map in order to determine its relative position to a goal location. The 180$^{\circ}$ laser scan range data is divided into 15 bins and concatenated to the robots angle and distance-to-goal, and the previously executed linear and angular velocity. This 19-dimensional tensor represents the state input $s_{t}$ to the policy. The prior controller takes as input the entire 180$^{\circ}$ laser scan data and angle-to-goal data in order to build its local potential field.

\subsubsection*{\textbf{Prior Controller}}
 For the prior controller, we utilise a variant of the Artificial Potential Fields controller introduced by Warren \textit{et al.} \cite{warren1989global}. It demonstrates a competent level of obstacle avoidance capabilities while exhibiting the same limitations faced by most reactive planners. As the standard form of this prior produces a deterministic action for the linear and angular velocities $[v, \omega]$, we approximate the distribution over these actions using Monte Carlo sampling. Given a known noise model of the laser scan range values $\mathcal{N}(0, \sigma^{2})$, we sample a 180-dimensional noise vector and add it to the laser scan value at a given state. This is passed through the controller producing the resulting linear and angular velocity. This is repeated \textit{N} times and the mean and variance of these values are computed to represent the distribution over the prior actions. At training, the distribution over the prior action space primarily serves as a medium for Gaussian exploration, allowing the policy to explore the surrounding state-action pairs for potential improvements, which we found important to stabilise training. We set this to a value of 0.3 throughout training.

\subsubsection*{\textbf{Policy}}
For training, we utilise Soft Actor-Critic (SAC) \cite{haarnoja2018soft}, an off-policy RL algorithm which naturally expresses its output as a distribution over its action space. SAC is known for its stability and robustness to hyper-parameters during training when compared to other off-policy algorithms. It incorporates an entropy regularisation term during training and is optimised to maximise a trade-off between expected return and entropy. This encourages exploration and prevents the policy from prematurely converging to a bad local optimum. The outputs from this policy follow the same convention outlined in section \ref{sec:mcf}. We utilise the implementation provided by OpenAI SpinningUp \cite{SpinningUp2018}.
 
\subsection{Evaluation of Training Performance}
We provide an evaluation of the training performance of our approach when compared to 3 learned baseline alternatives described below:\\
\textbf{End-to-end:} We train an agent using the standard Gaussian exploration provided in the SAC implementation.\\
\textbf{Baseline:} This method illustrates the naive use of demonstrations in the replay buffer. The replay buffer is filled with 50\% demonstrations from our prior and 50\% experience gathered by sampling the agent's policy.\\
\textbf{MCF (no-gating):} A variant of our proposed approach which does not rely on a gating function.\\

\vspace{-0.15cm}
\begin{figure}[t]
  \centering
  \includegraphics[width=0.5\textwidth]{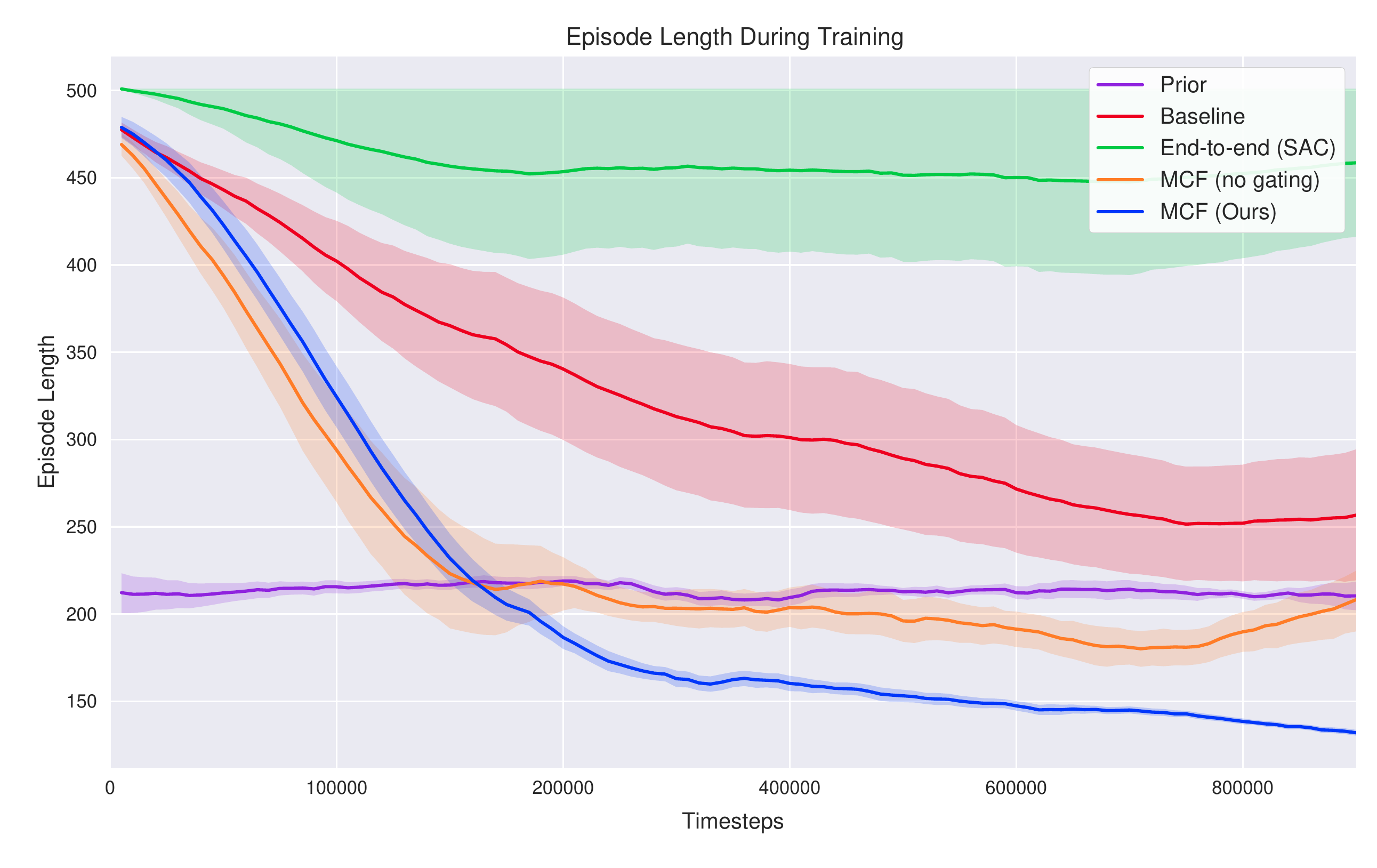}
  \vspace{-0.7cm}
  \caption{Learning curves showing average path length during training. MCF achieved the best performance compared to all alternatives with the least variance across 10 different seeds.}
  \label{training}
  \vspace{-0.15cm}
\end{figure}

\begin{figure}[t]
  \centering
  \includegraphics[width=0.49\textwidth]{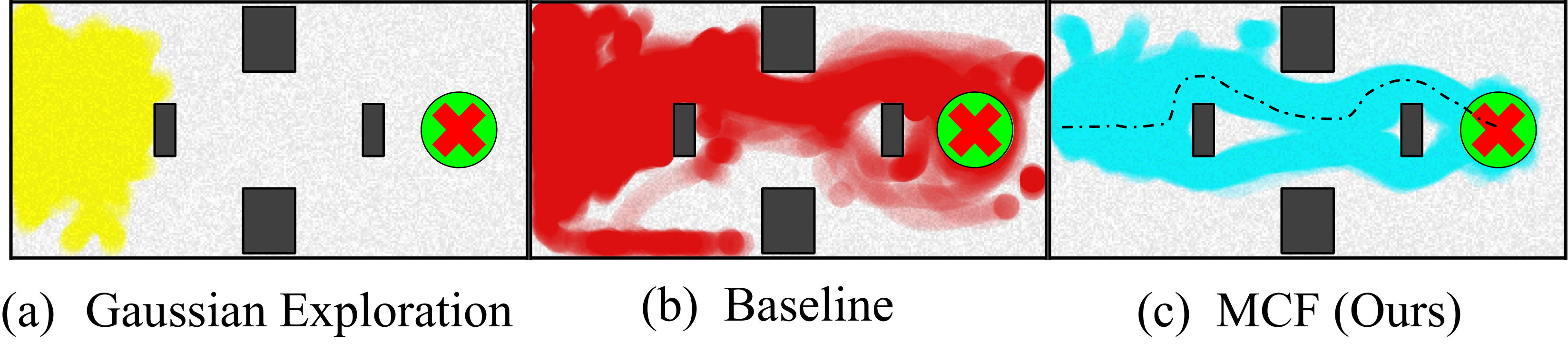}
  \vspace{-0.5cm}
  \caption{State space coverage during exploration. The dashed line in (c) illustrates the deterministic path taken by the prior controller. Note how our formulation explores the immediate surrounding regions of this demonstration.}
  \vspace{0.1cm}
  \label{explore}
  
\end{figure}

After every 5 episodes, we evaluated the policies' performance and the results are shown in Figure \ref{training}. All agents were trained across 10 different seeds. We additionally overlay the performance of the prior controller for comparison. Our approach shows the fastest convergence to an optimal policy and the least variance across all seeds. Note that MCF improves beyond the performance of the prior controller attaining a lower path length on average. The end-to-end based approach is shown to exhibit the worst performance with very high variance. The baseline approach also shows very high variance and is shown to converge to a suboptimal policy. We note that the no-gating variant of MCF, while showing lower variance, quickly convergences towards a suboptimal policy with similar performance to the prior. This is a result of it not being able to fully exploit its own policy and identify improvements beyond that of the prior, highlighting the importance of the gating parameter.\\
Figure \ref{explore} shows the state space coverage during exploration by standard Gaussian exploration, the baseline approach and MCF in an environment with a fixed start and goal location. It illustrates the poor performance of standard Gaussian exploration in sparse long horizon reward settings incapable of moving far beyond its initial position. The baseline, while benefiting from the demonstrations, is seen to spend time exploring unnecessary regions of the state space. MCF, on the other hand, illustrates structured exploration around the deterministic path of the prior controller (indicated by the dashed line), allowing it to focus on parts of the start space most relevant to the task while exploring the surrounding state-action regions for potential improvements. 

\begin{figure}[t]
  \centering
  \includegraphics[width=0.49\textwidth]{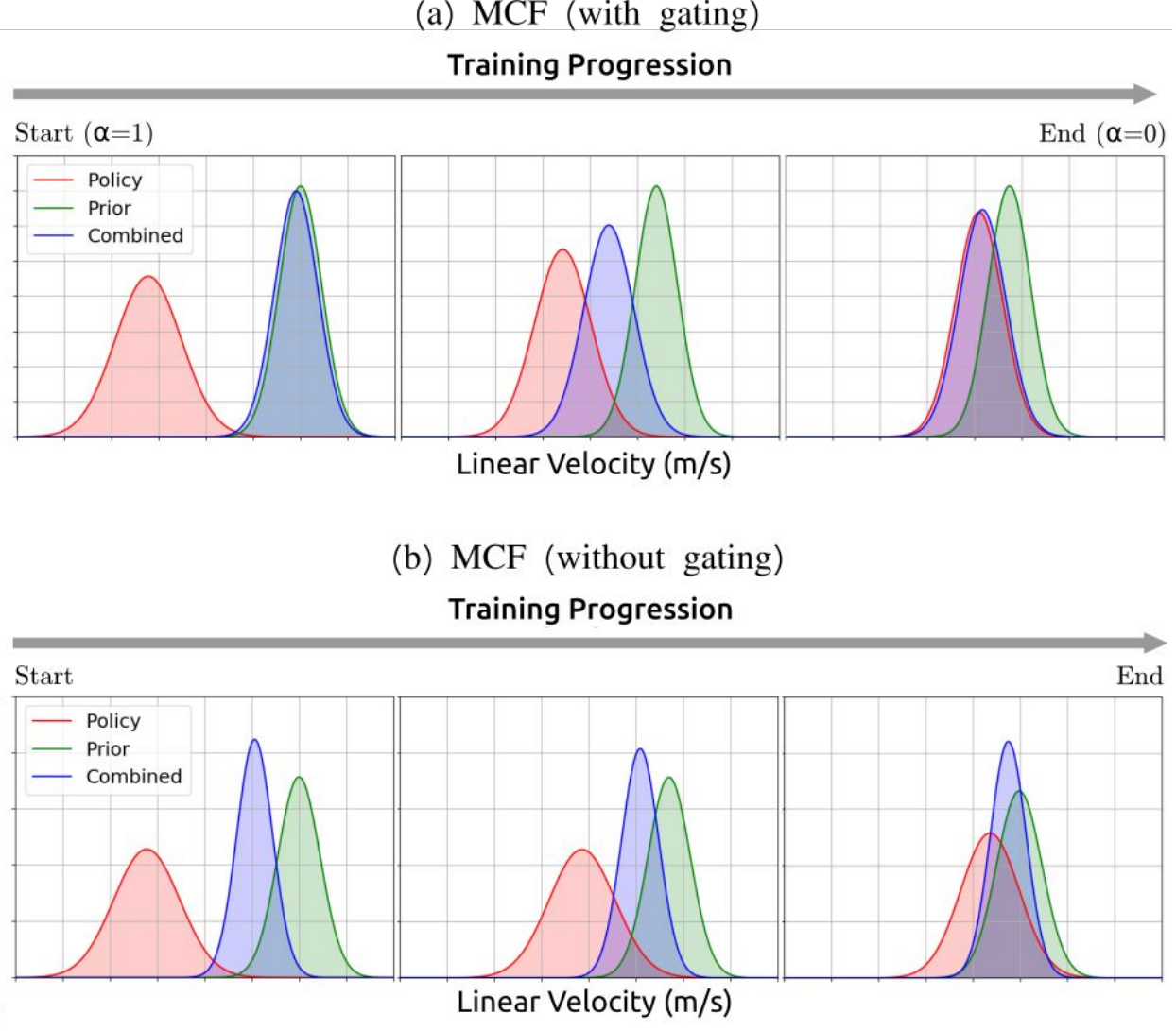}
  \vspace{-0.5cm}
  \caption{Progression of MCF during training showing the impact of the gating parameter over the course of training. We utilise a reverse logistic function which ranges from 1 to 0 in this work.}
  \vspace{-0.45cm}
  \label{gating}
\end{figure}

Figure \ref{gating} shows the progression of the composite MCF distribution used for exploration during training and the impact of the gating function. Without gating, the standard multiplicative fusion will result in a distribution which sits between the two systems. This limits the amount of guided exploration the prior can provide and the potential of the policy to fully exploit its own distribution in order to correct for extrapolation errors and improve beyond the prior. On the other hand, the gated variant allows us to sample actions from the distribution around the prior early on during training, allowing all the state-action regions surrounding the prior's trajectories to be updated. As the training progresses and the policy becomes more capable, we see its distribution naturally move closer towards that of the prior. Simultaneously, the gating function gradually shifts the resulting distribution to be fully on-policy. This allows the policy to correct any errors in its Q values while enabling it to explore potential improvements beyond that of the prior. Our results show that MCF constantly biases the policy's action distribution to be close to that of the prior with the necessary adjustments to overcome the inefficiencies of the prior.

\subsection{Evaluation of Deployed Systems}

\begin{table}[t]
\caption{Evaluation in Simulation Environment}
\label{tab:sim}
\begin{tabular}{@{}lcc
>{\columncolor[HTML]{C0C0C0}}c 
>{\columncolor[HTML]{C0C0C0}}c 
>{\columncolor[HTML]{FFFFFF}}l @{}}
\toprule
\multicolumn{1}{c}{} &
  \multicolumn{2}{c}{Training Environment} &
  \multicolumn{2}{c}{\cellcolor[HTML]{C0C0C0}Unseen Environment} &
   \\ \midrule
\textbf{Method} &
  \textbf{SPL} &
  \textbf{\begin{tabular}[c]{@{}c@{}}Actuation Time\\ (Steps)\end{tabular}} &
  SPL &
  \textbf{\begin{tabular}[c]{@{}c@{}}Actuation Time \\ (Steps)\end{tabular}} &
   \\ \midrule
Prior               & 0.793          & 207          & 0.666          & 305          &  \\
Policy Only         & 0.946          & 126          & 0.608          & 247          &  \\
\textbf{MCF (Ours)} & \textbf{0.965} & \textbf{119} & \textbf{0.728} & \textbf{227} &  \\
Random              & 0              & 500          & 0.148          & 478          &  \\ \bottomrule
\end{tabular}
\vspace{-0.45cm}
\end{table}

\begin{figure*}[t]
  \centering
  \includegraphics[width=\textwidth]{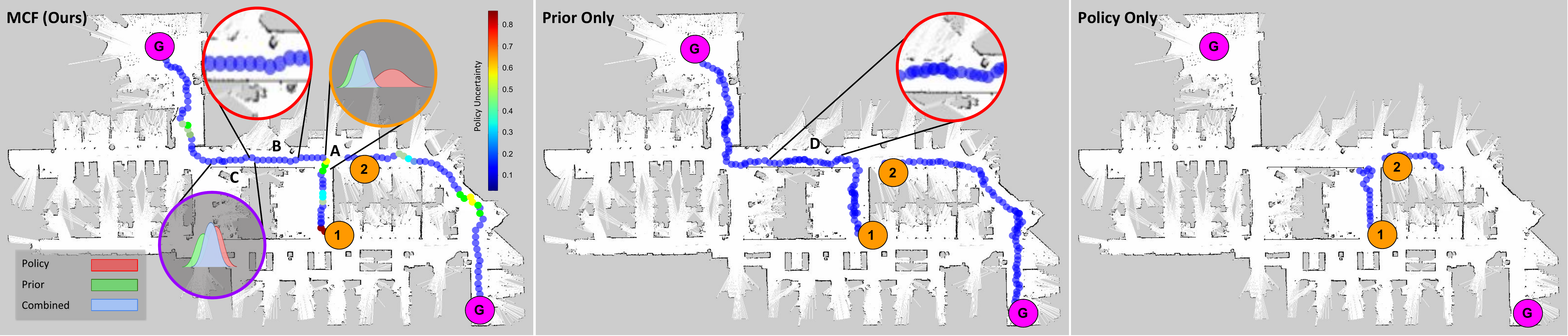}
  \vspace{-0.2cm}
  \caption{Trajectories taken by the real robot for different start (orange) and goal locations in a cluttered office environment with long narrow corridors. The trajectory was considered unsuccessful if a collision occurred. The trajectory taken by MCF is colour coded to represent the uncertainty in the linear velocity of the trained policy. We illustrate the behaviour of the fused distributions at key areas along the trajectory.}
  \label{maps}
  \vspace{-0.3cm}
\end{figure*}

We compare our deployment strategy to 3 different approaches in order to highlight the key advantages MCF exhibits during execution:\\
\textbf{Policy-only:} An individual policy trained end-to-end using SAC. Given that the standard Gaussian exploration used in the algorithm was insufficient to learn in the sparse reward setting, this algorithm was trained using Algorithm \ref{algorithm1}.\\
\textbf{Prior:} This represents the analytically derived reactive navigation controller based on the Artificial Potential Fields approach \cite{warren1989global}.\\
\textbf{ROS Move-Base:} For the real-world experiments, we compare to the state of the art classical local planner provided in the ROS navigation stack.\\
\textbf{Random:} Actions are randomly sampled from a uniform distribution between -1 and 1.\\

To evaluate the performance of these systems, we report the average Success weighted by (normalised inverse) Path Length (SPL) \cite{anderson2018evaluation} and the episodic actuation time. SPL weighs success by how efficiently the agent navigated to the goal relative to the shortest path. The metric requires a measure of the shortest path to goal which we approximate using the path found by an A-Star search across a 2000 $\times$ 1000 grid. An episode is deemed successful when the robot arrives within 0.2m of the goal. The episode is timed out after 500 steps and is considered unsuccessful thereafter. We do not report the SPL metric for the real robot experiments as we did not have access to an optimal path. We, however, provide distance travelled along each path and compare them to the distance travelled by a fine-tuned ROS \textit{movebase} planner. For computational efficiency, we utilised an ensemble of 5 trained policies to compute the policy distribution required by MCF across all evaluation runs.  In most states, we found that the prior controller utilised in this work exhibited little variance in the magnitude of laser noise present in the robot's laser scanner. As a result, we set the variance of the distribution to $max(\sigma^{2}_{mc}, C)$ where $\sigma^{2}_{mc}$ represents the variance computed by the Monte Carlo sampling approach and \textit{C} was empirically set to 0.2. This prevented the prior distribution from collapsing to a very confident value, limiting the impact of the policy on the overall system. We leave \textit{C} as a tuning variable which governs how risk-averse the system is allowed to behave in the real world.

\subsubsection{Simulation Environment Evaluation} 
We deploy the agent in both its training environment and an unseen environment to evaluate its performance when presented with unknown states. Table \ref{tab:sim} summarises the results. As expected, both our approach and the policy-only systems show superior performance in the training environment given that the policies have generalised well to all given states present, as indicated by the high SPL values. Additionally, we note that both learning-based systems exhibit lower actuation times than the prior, illustrating the efficiency gained via interaction. We attribute the higher actuation time of the prior to its oscillatory behaviour and lower SPL in cases where it got stuck in local minima. In the unseen environment, we see the key benefit of our approach which yields a higher SPL than both the prior and policy-only system and lowest actuation time. MCF attains the efficiency of the learned system while achieving a higher success rate than the policy-only system as a result of the prior fallback, which allows the agent to progress through regions that the policy would have otherwise failed.

\subsubsection{Real Robot Evaluation}
Given the close correspondence of laser scans between the simulation and training environment, we directly transfer the systems to a real robot. We utilise a PatrolBot mobile base shown in Figure \ref{front} which is equipped with a 180$^{\circ}$ laser scanner. All velocity outputs are scaled to a maximum of 0.25 \textit{m/s} before execution on the robot. The environment in which the system was deployed was a cluttered indoor office space which had been previously mapped using the laser scanner. We utilise the ROS \textit{ACML} package to localise the robot within this map to extract the necessary system inputs for the policy network and prior. Despite having a global map, the agent is only provided with global pose information with no additional information about its operational space. The environment also contained clutter which was unaccounted for in the mapping process. To enable large traversals through the office space, we utilise a global planner to generate target locations, 3 meters apart for our reactive agents to navigate towards. 

We evaluate the performance of the system on two different trajectories indicated as Trajectory 1 and Trajectory 2 in Table \ref{tab:real_robot} and Figure \ref{maps}. Trajectory 1 consisted of a lab space with multiple obstacles, tight turns, and dynamic human subjects along the trajectory, while Trajectory 2 consisted of narrow corridors never seen by the robot during training. As a comparative baseline, we include the performance achieved by a fine-tuned ROS \textit{move-base} planner. We summarise the results in Table \ref{tab:real_robot}. In all cases, the policy-only approach failed to complete the task without any collisions, exhibiting random reversing behaviours. We attribute this to its poor generalisation to certain states given the limited simulation training environment. The prior was capable of completing all trajectories however had the largest execution times as a result of its inefficient oscillatory behaviour. MCF was successful in all cases and showed significant improvements in efficiency when compared to the prior. We attribute this to the impact the policy has on the system. It also demonstrates competitive results with a fine-tuned move-base planner with similar distance coverage.\\ 
To gain a better understanding of the reasons for MCF's success when compared to the prior and policy only alternatives, we overlay the trajectories taken by these systems as shown in Figure \ref{maps}. The trajectory taken by MCF is colour-coded to illustrate the policy uncertainty in the linear velocity as given by the standard deviation of the policy ensemble outputs. We draw the readers attention to the region marked \textbf{A} which exhibits higher values of policy uncertainty. The multiplicative combination of the distributions at this region is shown within the orange ring. As expected, given the higher policy uncertainty at this point, the resulting composite distribution was biased more towards the prior which displayed greater certainty, allowing the robot to progress beyond this point safely. We note here that this is the particular region that the policy-only system failed as shown in Figure \ref{maps}. The purple ring at region C illustrates regions of low policy uncertainty with the composite distribution biased closer towards the policy. Comparing the performance benefit over the prior, we draw the readers attention to regions B and D which show the path profile taken by the agents. The dense darker path shown by the prior indicates regions of high oscillatory behaviour and significant time spent at a given location. On the other hand, we see that MCF does not exhibit this and attains a smoother trajectory which we can attribute to the policy having higher precedence in these regions, stabilising the oscillatory effects of the prior. We provide a video illustrating these behaviours with the real robot on our project page \footnote{\url{https://sites.google.com/view/mcf-nav/home}}.

\begin{table}[]
\centering
\caption{Evaluation for real world navigation}
\label{tab:real_robot}
\resizebox{0.49\textwidth}{!}{%
\begin{tabular}{@{}lcccc@{}}
\toprule
 & \multicolumn{2}{c}{\textbf{Trajectory 1}} & \multicolumn{2}{c}{\textbf{Trajectory 2}} \\ \midrule
\multicolumn{1}{c}{\textbf{Method}} & \textbf{\begin{tabular}[c]{@{}c@{}}Distance \\ Travelled\\ (meters)\end{tabular}} & \textbf{\begin{tabular}[c]{@{}c@{}}Actuation Time\\ (seconds)\end{tabular}} & \textbf{\begin{tabular}[c]{@{}c@{}}Distance \\ Travelled\\ (meters)\end{tabular}} & \textbf{\begin{tabular}[c]{@{}c@{}}Actuation Time\\ (seconds)\end{tabular}} \\ \midrule
Prior Only          & 34.398        & 271.1         & 24.8              & 148\\
End-to-end          & Fail          & Fail          & Fail              & Fail \\
\textbf{MCF}        & \textbf{32.9} & \textbf{184}  & \textbf{23.8}     & \textbf{131} \\
Move Base           & 33.6          & 154.1         & 23.3              & 153 \\
 \bottomrule
\end{tabular}%
}
\vspace{-0.30cm}
\end{table}

\addtolength{\textheight}{-0.1cm}

\section{Conclusions}
In this paper, we propose Multiplicative Controller Fusion (MCF), a stochastic fusion strategy for continuous control tasks. It provides a means to leverage the large body of work from the robotics community into learning-based approaches. MCF operates both during training and deployment. At training, we show that our gated formulation allows for low variance sample efficient learning from sparse, long-horizon reward tasks. At deployment, we demonstrate how MCF attains the efficiencies of learned policies in familiar states while falling back to a classical controller in cases of high policy uncertainty. This allows for superior performance when transferring a policy from simulation to the real world when compared to both the policy and classical systems individually. A limitation of our approach is when both the prior and policy are confident on totally different actions, which stagnates the resulting distribution. This may occur if the policy exploits an unwanted behaviour during training and hence acts considerably different from the prior. One way to mitigate this stagnation is to always fall back to the reliable classical controller in cases of total disagreement. Additionally, the gating function could be defined to directly relate to the policy's performance. We leave the exploration of these ideas to future work.

\section*{ACKNOWLEDGEMENTS}
The authors would like to thank Jake Bruce, Robert Lee, Mingda Xu, Dimity Miller and Jordan Erskine for their valuable and insightful discussions towards this contribution.

\bibliographystyle{IEEEtran}
\bibliography{IEEEabrv,references.bib,manual.bib}

\end{document}